\documentclass[10pt,twocolumn,letterpaper]{article}

\usepackage[pagenumbers]{cvpr} 

\definecolor{cvprblue}{rgb}{0.21,0.49,0.74}
\usepackage[pagebackref,breaklinks,colorlinks,allcolors=cvprblue]{hyperref}

\usepackage{wrapfig}
\usepackage{float}
\usepackage{booktabs}
\usepackage{array}
\usepackage[table]{xcolor}
\usepackage{adjustbox}
\usepackage{colortbl}
\usepackage{enumitem}
\usepackage{amsmath}

\definecolor{mygreen}{HTML}{59A14F}
\definecolor{myorange}{HTML}{DE7061}

\def\paperID{****} 
\def\confName{CVPR}
\def\confYear{2026}

\title{SUPER-AD: Semantic Uncertainty-aware Planning\\for End-to-End Robust Autonomous Driving}

\author{Wonjeong Ryu\textsuperscript{1}, Seungjun Yu\textsuperscript{1}, Seokha Moon\textsuperscript{2}, Hojun Choi\textsuperscript{1}, \\ Junsung Park\textsuperscript{1},
Jinkyu Kim,\textsuperscript{2} and Hyunjung Shim\textsuperscript{1 } \\
\textsuperscript{1}KAIST AI \qquad 
\textsuperscript{2}Korea University \\
}

\begin{document}
\maketitle
\begin{abstract}
End-to-End (E2E) planning has become a powerful paradigm for autonomous driving, yet current systems remain fundamentally uncertainty-blind.
They assume perception outputs are fully reliable, even in ambiguous or poorly observed scenes, leaving the planner without an explicit measure of uncertainty.
To address this limitation, we propose a camera-only E2E framework that estimates aleatoric uncertainty directly in BEV space and incorporates it into planning.
Our method produces a dense, uncertainty-aware drivability map that captures both semantic structure and geometric layout at pixel-level resolution.
To further promote safe and rule-compliant behavior, we introduce a lane-following regularization that encodes lane structure and traffic norms.
This prior stabilizes trajectory planning under normal conditions while preserving the flexibility needed for maneuvers such as overtaking or lane changes.
Together, these components enable robust and interpretable trajectory planning, even under challenging uncertainty conditions.
Evaluated on the NAVSIM benchmark, our method achieves state-of-the-art performance, delivering substantial gains on both the challenging NAVHARD and NAVSAFE subsets.
These results demonstrate that our principled aleatoric uncertainty modeling combined with driving priors significantly advances the safety and reliability of camera-only E2E autonomous driving.
\end{abstract}    
\section{Introduction}
\label{sec:intro}
\begin{figure*}[t]
\centering
\begin{subfigure}[t]{0.65\linewidth}
    \centering
    \includegraphics[width=\linewidth]{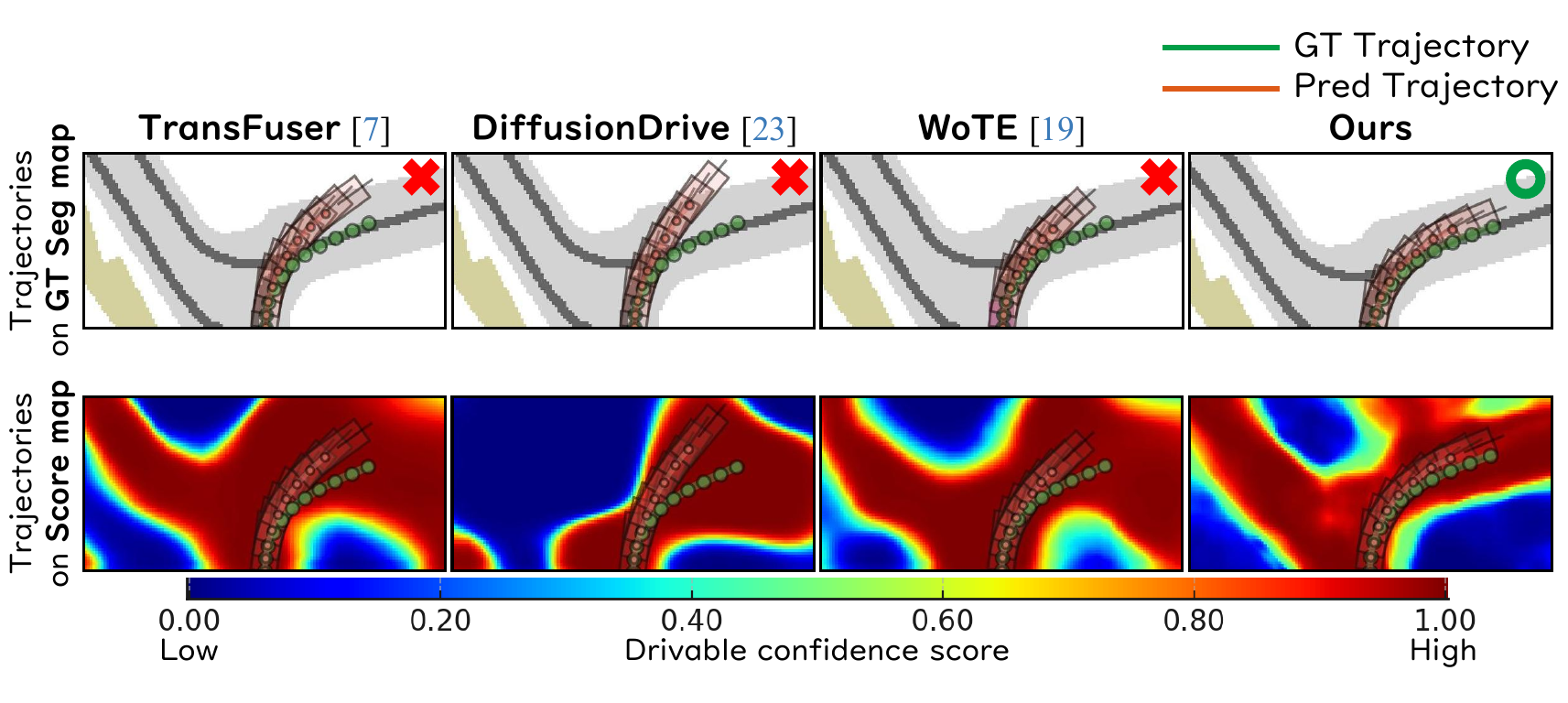}
    \vspace{-1.5em}
    \caption{} 
    \label{fig:teaser_ece:a}
\end{subfigure}\hfill
\begin{subfigure}[t]{0.33\linewidth}
    \centering
    \includegraphics[width=\linewidth]{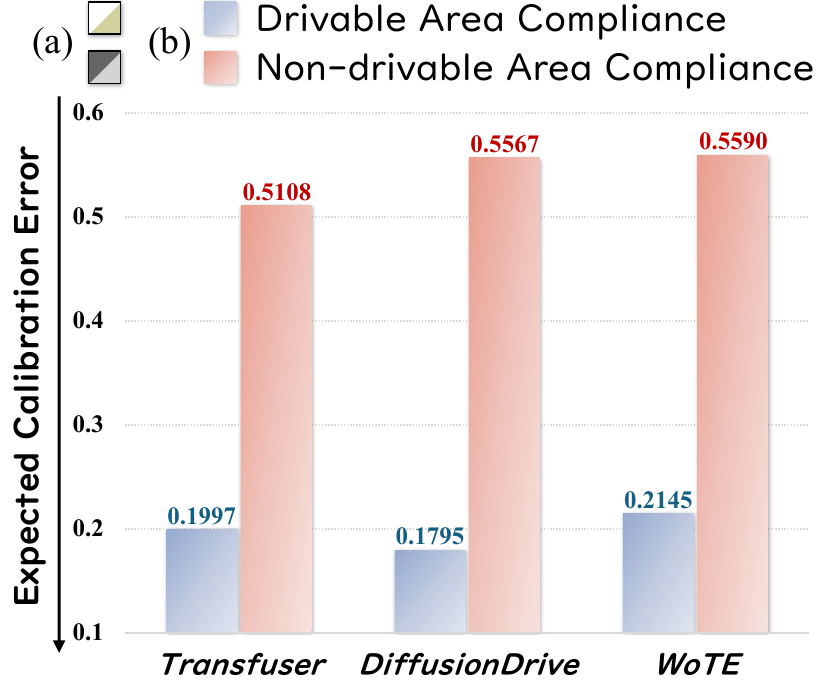}
    \vspace{-1.5em}
    \caption{} 
    \label{fig:teaser_ece:b}
\end{subfigure}

\vspace{-0.5em}
\caption{Illustration of overconfidence‑induced misplanning. It shows that the model deviates from the correct path toward non-drivable regions, as confirmed by the drivable confidence score map where these areas receive undesirably high confidence, leading to overconfident misplanning. (a) shows representative examples of such failure cases, where the model is misled by overconfident predictions. (b) presents quantitative evidence capturing this general trend.}

\vspace{-1.0em}
\label{fig:teaser_ece}
\end{figure*}

End-to-End (E2E) planning has emerged as a scalable and data-driven paradigm for autonomous driving~\citep{10614862}. Unlike conventional pipelines, which train perception, prediction, and planning as disjoint modules, E2E systems learn these components jointly within a single differentiable architecture. This unified design enables direct mapping from raw sensor observations to driving trajectories, greatly reducing manual engineering effort and unlocking richer driving strategies through holistic optimization. 

Despite its promise, current E2E systems exhibit a critical structural weakness: they treat perceptual outputs as fully reliable and ignore the underlying uncertainty. When the perception module encounters uncertain scenarios (e.g., ambiguous, dynamic, or poorly observed scenes), the planner has no mechanism to recognize or compensate for this uncertainty. This can lead to unsafe decisions~\citep{xu2014motion,seo2023safe,cao2025non}, such as entering non-drivable regions.~(Fig.~\ref{fig:teaser_ece:a}). This issue is reflected in the high expected calibration errors observed in E2E systems, indicating a persistent mismatch between confidence and accuracy~(Fig.~\ref{fig:teaser_ece:b}). Such miscalibration exposes a fundamental limitation of E2E planning and highlights the importance of explicit uncertainty modeling to achieve safer and more reliable decision-making.

Recently, several approaches have attempted to incorporate uncertainty into planning. Sensor fusion methods~\citep{chen2024vadv2,Chitta2023transfuser,li2024hydra,wote,yuan2024drama,diffusiondrive} combine camera and LiDAR to reduce observational noise and improve 3D understanding. 
However, they still overlook the uncertainty in perception itself, limiting their reliability during planning. 
Alternatively, HD map-based method~\citep{uncad} explicitly estimates both road boundary vectors and their confidence levels. While conceptually appealing, this method assumes that vector predictions are sufficiently accurate to support meaningful uncertainty estimation. This leads to a paradox: uncertainty estimates become unreliable precisely in regions where they are most needed~\citep{chen2025mapadsurvey}. Additionally, their vector-centered, ellipsoidal uncertainty representations are too rigid to capture the complex geometric and semantic richness of real-world driving scenes.

To address these limitations, we propose a novel E2E framework that explicitly integrates perceptual uncertainty into the trajectory planning process. Our model computes uncertainty by sampling from the predictive distribution of the output logits from the semantic segmentation head in BEV space. By measuring the variance across sampled probabilities, the system derives a principled Bayesian estimate of perceptual uncertainty. The resulting semantic predictions and uncertainty values are then fused into a high-resolution drivable score map. Each pixel in this map is categorized as drivable, non-drivable, or uncertain, with an associated confidence score. This dense pixel-level representation captures geometric layout in BEV and semantic context from segmentation outputs, enabling the planner to reason about uncertainty at pixel-level granularity. In contrast to prior works that treat uncertainty as a mere exclusion signal, our model leverages it as contextual guidance, enabling flexible, interpretable, and risk-aware trajectory planning in complex driving environments.

Although leveraging uncertainty enhances adaptability in complex scenarios, safe driving also depends on respecting road structure and traffic norms. E2E models learn directly from data, but essential norms such as lane following are often implicit and underrepresented. Consequently, these models often drift toward lane boundaries or perform unsafe deviations, even in areas with high confidence. To address this, we incorporate a lane-following regularization term that encodes lane keeping as a safety prior~\citep{2022LTP,wang2025empirical}. This term promotes alignment with the lane center under normal conditions, while still allowing necessary deviations, such as during overtaking or lane changes. By translating intuitive driving norms into explicit learning constraints, our approach produces trajectories that are both uncertainty-aware and aligned with safe-driving practices. This enhancement moves beyond simple avoidance of uncertain areas, yielding robust, safe, and human-aligned driving behaviors across diverse and challenging scenes.

We validate our approach on the NAVSIM benchmark \citep{Dauner2024NEURIPS,Cao2025CORL}, which features challenging real-world scenes under a non-reactive evaluation protocol. Our method achieves state-of-the-art performance, significantly outperforming prior works in both the NAVHARD \cite{Cao2025CORL} and NAVSAFE \cite{sima2025centaur} subsets, which emphasize robustness and safety. These results empirically demonstrate that our camera-only method, which deeply integrates uncertainty understanding and lane-following priors, not only matches but surpasses sensor fusion baselines, offering a practical and safer alternative for large-scale autonomous driving deployment.
\newline
\newline
\noindent In summary: 
\begin{itemize}
    \item We model uncertainty in the Bird's-Eye View~(BEV) feature space.~This process generates high-resolution uncertainty-aware drivable score map to guide trajectory planning~(\cref{sec:uncertainty}).
    \item We introduce a lane-following regularization term to ensure trajectory stability. This term acts as a rigorous stabilizing prior for the driving policy~(\cref{sec:lane}).
    \item Our camera-only approach achieves state-of-the-art performance on the NAVSIM benchmark.~It shows substantial improvements on challenging~(NAVHARD) and safety-critical~(NAVSAFE) splits.
\end{itemize}

\section{Related Work}
\label{sec:formatting}
\begin{figure*}[!t]
\centering
\includegraphics[width=0.925\linewidth]{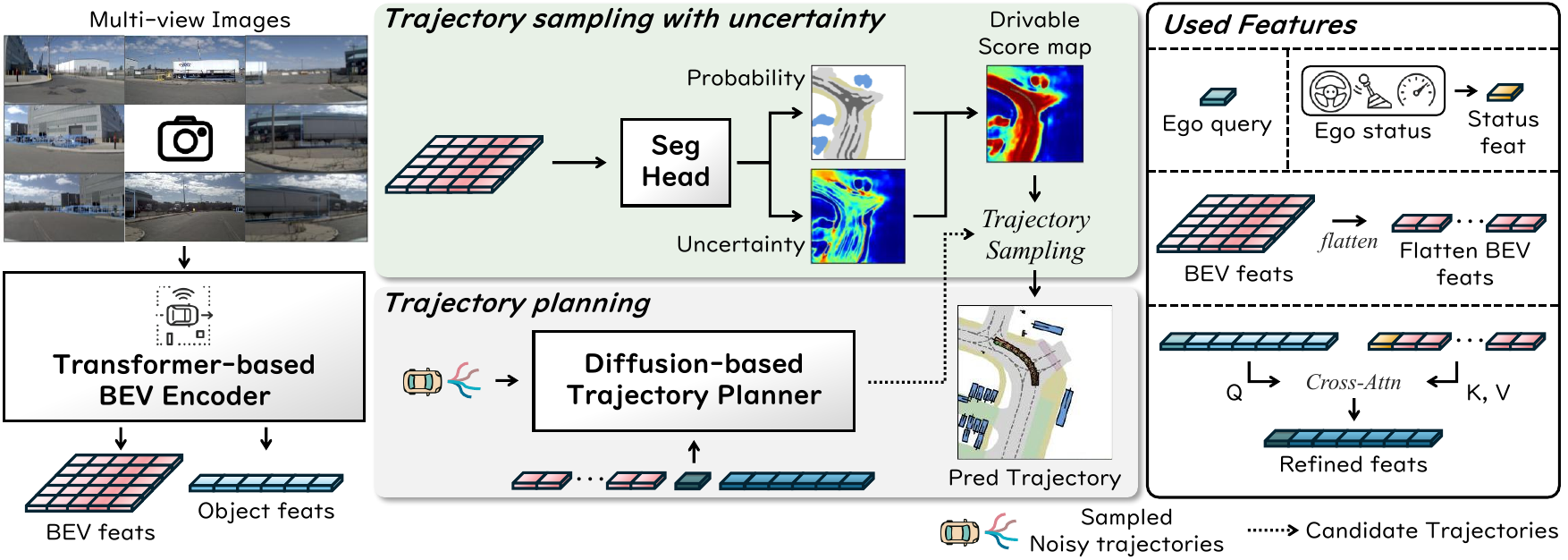}
\vspace{-0.5em}
\caption{
Overview of our framework at inference time. Our model first extracts Bird's-Eye View (BEV) features from multi-view images. The segmentation head then uses these features to predict a distribution composed of logits and uncertainty. In parallel, the planner based on DiffusionDrive~\citep{diffusiondrive} estimates candidate trajectories using the encoder outputs, and ego vehicle features. 
After estimate trajectories, predicted distribution is utilized to evaluate the candidates.
This weighting mechanism prevents trajectories from being sampled by penalizing those that pass through high uncertainty areas.
}
\vspace{-1.0em}
\label{fig:main_fig}
\end{figure*}

\subsection{End-to-End Autonomous Driving} 
End-to-End (E2E) autonomous driving has steadily advanced alongside developments in perception technologies such as Convolutional Neural Networks (CNNs) and Transformers. In particular, early methods stemmed from perception-based approaches that jointly addressed multiple tasks like detection and segmentation. For example, UniAD \citep{uniad} integrated perception tasks to improve planning performance, while Transfuser \citep{Chitta2023transfuser} demonstrated the feasibility of end-to-end planning by fusing LiDAR and camera inputs using a Transformer.

Subsequent research focused on scene representation and architectural efficiency.~VAD \citep{jiang2023vad} represented the scene through vectorization, which SparseDrive \citep{sparsedrive} extended into a sparse scene representation. DriveTransformer \citep{drivetransformer} proposed a Transformer-based parallelized architecture, and PARA-Drive \citep{paradrive} achieved state-of-the-art performance and improved runtime efficiency by co-training the perception, prediction, and planning modules. DRAMA \citep{yuan2024drama} introduced the State Space Model in place of Transformers, implementing more efficient feature fusion and trajectory generation.

More recently, there has been a diversification in modeling paradigms. DiffusionDrive \citep{diffusiondrive} demonstrated strong performance by leveraging diffusion policies, while LAW \citep{law} and WoTE \citep{wote} enhanced planning safety and interpretability using world model-based approaches. GraphAD \citep{graphad} strengthened the understanding of interactions between surrounding objects via a graph structure. Additionally, models like FASIONAD++ \citep{fasionad++}, OpenDriveVLA \citep{opendrivevla}, and Hydra-MDP \citep{li2024hydra} have expanded higher-order reasoning capabilities and generalization through VLM integration or multi-teacher knowledge distillation.

Despite this remarkable progress in end-to-end planning, existing methods still struggle to ensure safe and reliable driving in safety-critical scenarios.~Therefore, this study aims to address this challenge by leveraging uncertainty.

\subsection{Uncertainty in Autonomous Driving}
Uncertainty estimation, broadly used in serveral tasks \citep{segment,depth}, has been introduced into autonomous driving.
In line with this, one of the approaches \citep{producing} analyzes the influence of map uncertainty and emphasizes that explicitly considering map estimation uncertainty leads to more robust predictions.
Alternatively, MapDiffusion \citep{mapdiffusion} represents uncertainty by repeatedly predicting vectorized maps and computing the variance across samples to generate an uncertainty map.
Extending to depth level, GaussianLSS \citep{gaussianlss} explicitly models depth uncertainty by learning the mean and variance of depth distributions, thereby improving the reliability of image-based depth estimation.
In the trajectory level, CUQDS \citep{cuqds} proposed a framework that effectively mitigates predictive uncertainty under distribution shift conditions for trajectory prediction.
Finally, UncAD \citep{uncad} models uncertainty using a Laplace distribution and provides practical guidelines into E2E framework.

These recent studies have successfully modeled uncertainty in their respective ways.~Following this recent trend in uncertainty representation \citep{producing,uncad}, our work proposes a Bayesian learning–based approach that estimates high-resolution drivable confidence map across the entire BEV feature space \citep{NIPS2017_uncertainty}, which explicitly incorporates uncertainty into the planning process.~Although vector map–based methods are also effective, their uncertainty estimation remains highly dependent on the accuracy of initial vector predictions.
In contrast, our approach operates over the entire feature space, enabling more reliable uncertainty quantification that contributes to planning safer trajectories.
\section{Method}
While LiDAR offers precise 3D geometry, its cost and hardware complexity hinder scalability.
As a more scalable, camera-only approaches have emerged as a practical alternative, and recent transformer-based models~\citep{li2022bevformer,bevformerv2,streampetr} convert multi-view images directly into Bird’s-Eye-View (BEV) features, enabling strong 3D representations.

However, camera-based models inherently face higher perceptual uncertainty, which spans geometric and semantic uncertainties from raw sensor data.
This perceptual uncertainty becomes pronounced under occlusions caused by objects or adverse weather and at long ranges, where elevated sensor noise significantly degrades the reliability of the measurement.
In our method, we capture this perceptual uncertainty through the BEV segmentation head.
Although the model predicts pixel-level semantic labels, each BEV pixel is anchored to a specific location and thus implicitly carries geometric information.
We therefore treat the resulting semantic uncertainty on BEV pixels as a form of perceptual uncertainty.

To mitigate the impact of this perceptual uncertainty, we propose a transformer-based BEV encoder that models perceptual uncertainty directly in the BEV map segmentation and propagates it to the planning.
The perception module performs BEV map segmentation to represent the driving scene and simultaneously conducts 3D object detection to refine the geometric understanding of objects.
Notably, the segmentation head outputs both per-pixel classes and corresponding uncertainty estimates, enabling the network to recognize uncertain, ambiguous regions and mitigate overconfidence in map prediction.
These outputs are consolidated into an uncertainty-aware drivable score map.
Segmentation classes are grouped into drivable and non-drivable sets, and each pixel is assigned a drivable score that quantifies the reliability of its assignment. 
As this uncertainty is computed in the perception module, our method is planner-agnostic and can be seamlessly integrated into diverse planning modules.

Our planning module, built upon the DiffusionDrive \citep{diffusiondrive} decoder, receives BEV features along with ego-vehicle and agent features to generate multiple candidate trajectories.
Each candidate trajectory is then assigned a likelihood from the uncertainty-aware drivable score map, which serves as a weight that penalizes paths traversing uncertain or non-drivable regions.
To prevent the planner from being biased toward low uncertainty regions at the expense of ignoring lane structures, we introduce a lane-following regularization that encourages alignment with the lane center.
As a result, our model produces robust and safe trajectories even under uncertain scenarios, significantly reducing safety-critical failures.

\subsection{Uncertainty Model}
\label{sec:uncertainty}
In this section, we describe constructing an uncertainty-aware drivable score map from the perceptual uncertainty to enable risk-aware trajectory planning.
We model uncertainty densely across the entire BEV feature space because prior vector-map approaches represent it only at sparse boundary points. This design allows the planner to generate safe trajectories not only about clearly drivable or non-drivable areas but also about ambiguous, uncertain regions where the environment cannot be confidently perceived. By providing this richer (denser) uncertainty representation, the planner can make safer and more informed trajectory decisions in complex scenes.

To formally model this uncertainty based on conventional method~\citep{NIPS2017_uncertainty}~(See supplementary materials), we treat each BEV pixel's class logits as a Gaussian distribution with a mean $\boldsymbol{\mu}\in\mathbb{R}^{H\times W\times K}$ and a standard deviation $\boldsymbol{\sigma}\in\mathbb{R}^{H\times W\times K}$. 
The $\boldsymbol{\mu}$ denotes the central estimate of the model prediction for each pixel, while the $\boldsymbol{\sigma}$ represents the class-wise logit aleatoric uncertainty caused by factors such as sensor noise or occlusions.
In this way, the model learns a distribution over predictions instead of producing a single deterministic output.
This enables the model to account for the inherent uncertainty present in the scene.
Accordingly, the training objective is defined as follows:
\vspace{-0.25em}
\begin{align}
\mathcal{L}_{\mathrm{perc}}
=&
- \log \left(
  \frac{1}{T}\sum_{t=1}^{T}
  \mathrm{softmax}\!\big(\,\boldsymbol{\mu} + \boldsymbol{\sigma}\odot\boldsymbol{\varepsilon}^{(t)}\,\big)_{y}
\right).
\end{align}
According to this formulation, $\mathbf{y}_i$ denotes the ground truth target for a given input $\mathbf{x}_i$, and $\boldsymbol{\varepsilon}^{(t)} \sim \mathcal{N}(\mathbf{0},\mathbf{I})$ denotes i.i.d. standard gaussian noise used to sample logits through a differentiable change of variables.
The loss approximates the expected softmax under the predicted gaussian over logits by sampling several logits from this distribution, applying the softmax to each, and averaging the resulting class probabilities.
Since this sampling process introduces stochasticity that could hinder differentiability, we reparameterize the random logits so that gradients propagate through the stochastic variables while uncertainty is preserved in the forward pass.

To propagate this perceptual uncertainty into the planning, we incorporate it during the process of sampling candidate trajectories.
We first model uncertainty in the logit space, where the class-wise $\boldsymbol{\sigma}$ is defined for logits rather than probabilities.
Its magnitude alone cannot be directly interpreted across classes, since even the same $\boldsymbol{\sigma}$ reflects different confidence levels depending on the corresponding $\boldsymbol{\mu}$.
To obtain a consistent probabilistic interpretation, the logit distribution is mapped into probability space, and the expected probabilities $\bar{p}$ are computed as:
\vspace{-0.25em}
\begin{equation}
\bar{p}
\;\approx\;
\frac{1}{T}\sum_{t=1}^{T}
\mathrm{softmax}\!\big(\,\boldsymbol{\mu} + \boldsymbol{\sigma}\odot\boldsymbol{\varepsilon}^{(t)}\,\big).
\end{equation}
\noindent The $\bar{p}$ computed via Monte Carlo sampling converts logit-level predictive distribution into probability space and provides uncertainty-aware class probability estimates that the planner exploits for trajectory evaluation.
Based on this probabilistic representation, the classes are grouped into drivable $\mathcal{P}_{drivable}$ and non-drivable categories:
\vspace{-0.25em}
\begin{align}
\bar{p}_{\mathrm{pos}} \hspace{0.9em}&= \sum_{c \in \mathcal{P}_{drivable}} \bar{p}_{c}, \\
H_{\mathrm{group}} &= -\,\bar{p}_{\mathrm{pos}}\log_{2}\!\big(\bar{p}_{\mathrm{pos}}\big)
-\big(1-\bar{p}_{\mathrm{pos}}\big)\log_{2}\!\big(1-\bar{p}_{\mathrm{pos}}\big). \notag
\end{align}
\noindent The group entropy $H_{\mathrm{group}}$ quantifies decision ambiguity using a binary cross-entropy formulation, normalized between 0 and 1.
A value of 0 indicates full confidence, while 1 corresponds to maximum uncertainty at $\bar{p}_{\mathrm{pos}} = 0.5$.
As a final step, we define a pixel‑level safety score $S_{safe}$ as follows:
\vspace{-0.25em}
\begin{equation}
S_{\mathrm{safe}}
=
\bar{p}_{\mathrm{pos}}\,(1 - H_{\mathrm{group}})
+ 0.5\, H_{\mathrm{group}}.
\end{equation}
\noindent As $H_{\mathrm{group}}$ decreases to 0, $S_{\mathrm{safe}}$ converges toward $\bar{p}_{\mathrm{pos}}$, meaning the planner fully trusts the predicted drivable probability.
Conversely, as $H_{\mathrm{group}}$ increases to 1, $S_{\mathrm{safe}}$ moves toward 0.5, so the planner takes a more cautious and neutral decision under high uncertainty.

For trajectory sampling, we project each candidate onto the uncertainty-aware drivable score map, defined by $S_{\mathrm{safe}}$ in BEV space, and assign a score based on the minimum safety value along its path.
The selection probability of each candidate is then weighted accordingly, and any path intersecting non-drivable pixels is discarded.
This process prefers to prioritize with high drivable probability and low ambiguity.
As a result, failure cases such as off-road driving are significantly reduced, demonstrating the enhanced capability of the model for safer and more reliable planning.

\subsection{Lane-Following Regularization}
\label{sec:lane}
Uncertainty-aware planning encourages the vehicle to stay within drivable regions.
However, such planning alone is insufficient to ensure that the resulting trajectories faithfully follow the intended lane structure.
To mitigate this issue, a constraint to the lane centerline typically stabilizes behavior~\citep{wang2025empirical}, yet legitimate maneuvers such as turns or lane changes require the vehicle to intentionally deviate from the centerline.
The planner, therefore, decides when to follow the lane center and apply the constraint only then.
We introduce a two-part loss that produces guides to the model on when to follow the lane center and enforces adherence to the centerline when necessary.

First, to learn the intent of following the lane, we define two masks.
The first, $M_{gt}$, indicates for each point on the expert trajectory whether it follows its lane centerline.
The second, $M_{pred}$, is the corresponding mask defined over the predicted trajectory points.
To account for spatial misalignment between predicted and expert trajectories, each predicted point $\hat{\mathbf{x}}_t$ is matched to its nearest expert point $\mathbf{x}^{gt}_{\pi(t)}$.
The intent loss is computed as the L1 difference between the masks after this nearest-neighbor association:
\vspace{-0.25em}
\begin{equation}
\mathcal{L}_{\text{intent}}
=\frac{1}{T_{\text{pred}}}\sum_{t=1}^{T_{\text{pred}}}
\bigl|\, M_{pred}(t)-M_{gt}\bigl(\pi(t)\bigr)\,\bigr|.
\end{equation}
Minimizing $\mathcal{L}_{\text{intent}}$ enables the model to learn when lane-following is appropriate.

Next, we enforce geometric adherence to the lane center only at predicted points for which the matched expert points require lane following.
Each such predicted point is projected into the BEV space, and we compute the pixel-level L2 distance to the nearest pixel labeled as the centerline class on the BEV segmentation map:
\begin{equation}
\mathcal{L}_{\text{center}}
=\frac{1}{N_c}\sum_{t=1}^{N_c}
\| \Pi(\hat{\mathbf{x}}_t) - \mathcal{C}_{\text{center}} \|_2,
\end{equation}
where $N_c$ is the number of predicted points required to follow the lane center, 
$\Pi(\cdot)$ denotes the projection into the BEV space, 
and $\mathcal{C}_{\text{center}}$ represents the set of pixels labeled as the lane center class on the BEV segmentation map.
This encourages those designated points to remain close to the lane centerline while avoiding penalties during legitimate deviations such as turns or lane changes.

Finally, the lane-following regularization combines both components as follows:
\begin{equation}
\mathcal{L}_{\text{lane}}
=\lambda_{\text{intent}}\,\mathcal{L}_{\text{intent}}
+\lambda_{\text{center}}\,\mathcal{L}_{\text{center}}.
\end{equation}
The model first learns when to follow the lane center, and the loss subsequently penalizes any deviation precisely at those moments.
This enhances both safety and robustness by acting as a stabilizing regularizer, especially by preventing the planner tendency to veer from its lane to avoid high uncertainty regions.

\subsection{Objective Function}
The BEV objective aims to train the representation to be both semantics-balanced and geometry-aware by coupling segmentation with 3D detection.
Specifically, the overall BEV loss $\mathcal{L}_{\text{BEV}}$ augments the perception loss with Focal and Dice terms to mitigate class imbalance and preserve fine-grained structures. 
In addition, a 3D object detection loss injects explicit geometric supervision, encouraging the BEV features to encode the 3D dynamic properties of objects.
The objective is defined as:
\begin{equation}
\mathcal{L}_{\text{BEV}} 
= \lambda_{\text{perc}} \, \mathcal{L}_{\text{perc}} 
+ \lambda_{\text{focal}} \, \mathcal{L}_{\text{focal}} 
+ \lambda_{\text{dice}} \, \mathcal{L}_{\text{dice}} 
+ \lambda_{\text{det}} \, \mathcal{L}_{\text{det}},
\end{equation}
where $\lambda_{\text{perc}}, \lambda_{\text{focal}}, \lambda_{\text{dice}},$ and $\lambda_{\text{det}}$ balance the contribution of each term, and the resulting semantics-balanced and geometry-aware BEV encoder provides structured evidence for the planner.

\begin{figure*}[t!]
\centering
\includegraphics[width=0.925\linewidth]{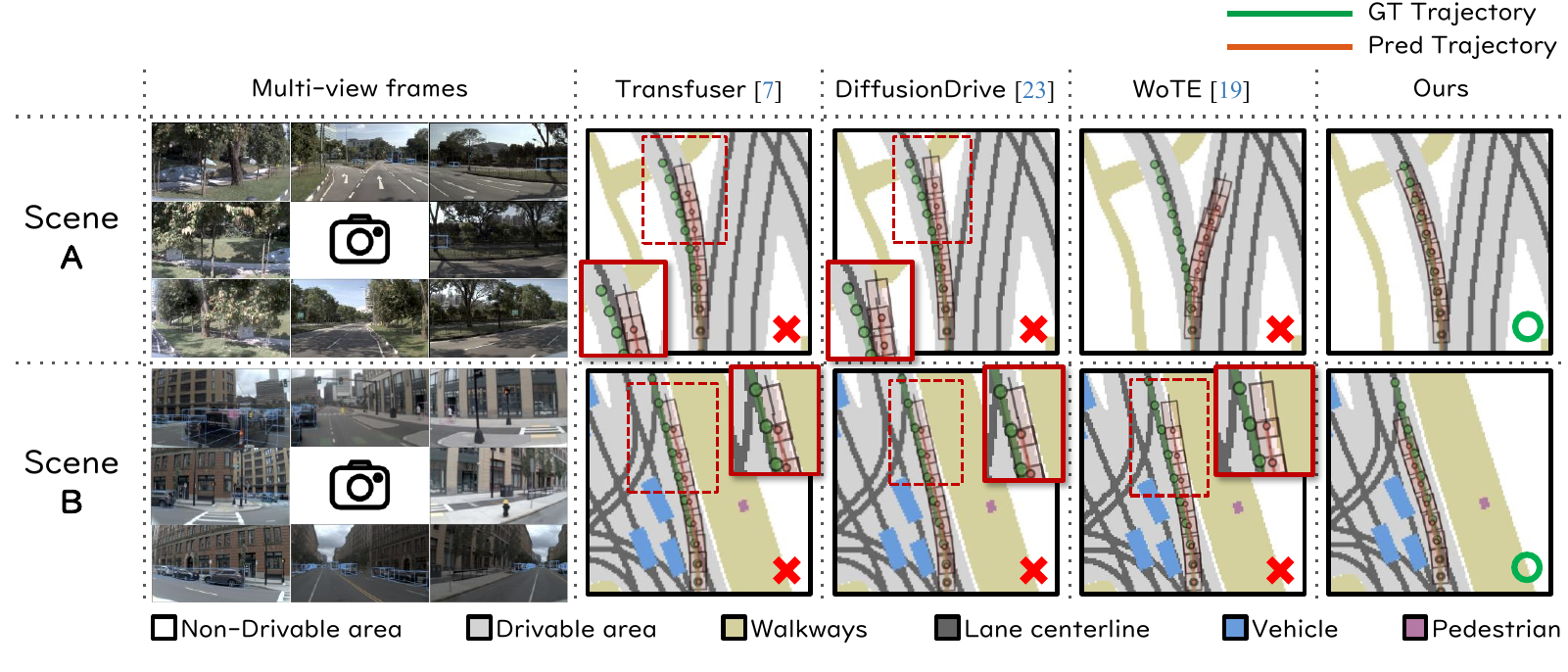}
\vspace{-0.5em}
\caption{
Qualitative results on challenging scenes. Scene \textbf{A} depicts a diverging lane entry where occlusions and viewpoint rotations increase uncertainty. Scene \textbf{B} involves safe lane changes in the presence of multiple objects, highlighting risk-aware planning.
\vspace{-1.0em}
}
\label{fig:qualitative_fig}
\end{figure*}

Building on this BEV feature, the planning module incorporates uncertainty into the trajectory selection process and aligns the highest scoring trajectory with the expert.
It scores each candidate trajectory by combining its class probabilities with the score map obtained in \cref{sec:uncertainty}, and the classification term encourages the expert consistent candidate to achieve the highest score.
At the same time, the ranking loss enforces the correct order among predicted scores.
After obtaining the highest scoring trajectory, the regression term minimizes the spatial gap between the selected and expert trajectories.
The planning objective is expressed as:
\begin{equation}
\mathcal{L}_{\text{planning}}
= \lambda_{\text{cls}} \, \mathcal{L}_{\text{cls}} 
+ \lambda_{\text{traj}} \, \mathcal{L}_{\text{traj}} 
+ \lambda_{\text{rank}} \, \mathcal{L}_{\text{rank}},
\end{equation}
where $\mathcal{L}_{\text{cls}}$, $\mathcal{L}_{\text{traj}}$, and $\mathcal{L}_{\text{rank}}$ denote the classification, regression, and ranking losses, respectively.

Finally, the overall objective combines all components, including the lane-following regularization described in \cref{sec:lane}:
\begin{equation}
\mathcal{L}_{\text{total}}
= \mathcal{L}_{\text{BEV}}
+ \mathcal{L}_{\text{lane}}
+ \mathcal{L}_{\text{planning}}.
\end{equation}
Through this unified objective, the planner learns to avoid non-drivable and even uncertain regions by considering uncertainty, while maintaining safe and robust lane-following behavior.

\section{Experiments}
\subsection{NAVSIM}
The NAVSIM dataset~\citep{Dauner2024NEURIPS,Cao2025CORL} enables End-to-End~(E2E) planning evaluation in real-world scenes using a non-reactive policy.
It is constructed from OpenScene~\citep{openscene2023} a 2Hz lightweight release of nuPlan~\citep{nuplan} after filtering out simple scenes. 
As a result, NAVSIM provides a rigorous and complex setting.
Although nuScenes~\citep{nuscenes2019} is another evaluation option, it is based on open-loop evaluation and predominantly contains simpler straight-forward scenes~\citep{li2024ego}, thus we adopt NAVSIM to demonstrate the effectiveness of our method.
Modeling uncertainty substantially improves safety, particularly under challenging scenarios.
Therefore, we provide further experimental evaluation using the NAVHARD~\citep{Cao2025CORL} and NAVSAFE~\citep{sima2025centaur} splits of the NAVSIM test set. \\

\noindent\textbf{NAVHARD \& NAVSAFE.} NAVHARD is a test split assembled for the NAVSIM leaderboard by collecting challenging scenes.
It includes situations such as unprotected turns and dense traffic.
At the same time, NAVSAFE is a safety‑focused benchmark subset proposed by \cite{sima2025centaur}.
To construct NAVSAFE, the benchmark follows the NHTSA crash-scenario definitions~\citep{najm2007precrash} and identifies the corresponding frames within NAVTEST~(the NAVSIM test split) via a semi-automatic pipeline.
Accordingly, to evaluate robustness under these challenging and safety‑critical scenes, we additionally report results on both NAVHARD and NAVSAFE.

\begin{table}[!t]
  \caption{Comparison of methods on PDM score (C: Camera, L: LiDAR). \textbf{Bold}: best; \underline{underline}: second best. Results are reported on the NAVTEST split. The compared architectures are based on a ResNet-34~\citep{he2016deep} backbone. The experiments were conducted using the NVIDIA RTX A6000.}
\vspace{-0.5em}
  \label{tab:pdms_v1}
  \begin{adjustbox}{width=\linewidth}
  \renewcommand{\arraystretch}{1.05}
  \large
  \setlength{\heavyrulewidth}{1.5pt}
  \setlength{\lightrulewidth}{0.5pt}
  \centering
  \begin{tabular}{@{}l|c|cc|ccc|>{\columncolor{gray!12}}c}
  
    \toprule
    Method & Modality & NC~$\uparrow$ & DAC~$\uparrow$ & EP~$\uparrow$ & TTC~$\uparrow$ & C~$\uparrow$ & PDMS~$\uparrow$ \\
    \midrule

    VAD-v2~\citep{chen2024vadv2} & C \& L & 97.2 & 89.1 & 76.0 & 91.6 & \textbf{100}  & 80.9 \\
    Transfuser~\citep{Chitta2023transfuser} & C \& L & 97.9 & 92.8 & 79.2 & 92.8 & \textbf{100}  & 84.0 \\
    DRAMA~\citep{yuan2024drama} & C \& L & 98.0 & 93.1 & 80.1 & \underline{94.8} & \textbf{100}  & 85.5 \\
    Hydra-MDP~\citep{li2024hydra} & C \& L & \underline{98.3} & 96.0 & 78.7 & 94.6 & \textbf{100}  & 86.5 \\
    DiffusionDrive~\citep{diffusiondrive} & C \& L & 98.2 & \underline{96.2} & \textbf{82.2} & 94.7 & \textbf{100}  & \underline{88.1} \\
    WoTE~\citep{wote} & C \& L & \textbf{98.5} & \textbf{96.8} & \underline{81.9} & \textbf{94.9} & 99.9 & \textbf{88.3} \\
    \noalign{\vskip 2pt} \hline\hline\noalign{\vskip 2.5pt} 

    UniAD~\citep{uniad} & C      & 97.8 & 91.9 & 78.8 & 92.9 & \textbf{100} & 83.4 \\
    PARA-Drive~\citep{paradrive} & C      & \underline{97.9} & 92.4 & 79.3 & \underline{93.0} & 99.8 & 84.0 \\
    LAW~\citep{law} & C      & 96.4 & \underline{95.4} & \textbf{81.7} & 88.7 & \underline{99.9} & \underline{84.6} \\
    \midrule
    Ours           & C      & \textbf{98.1} & \textbf{97.0} & \underline{81.5} & \textbf{93.9} & 99.8 & \textbf{87.7} \\

    \bottomrule
  \end{tabular}
\end{adjustbox}
\vspace{-1.0em}
\end{table}

\begin{table*}[!t]
  \caption{Comparison of methods on extended PDM score. (C: Camera, L: LiDAR). \textbf{Bold}: best; \underline{underline}: second best. Split (a) corresponds to the full NAVTEST, whereas split (b) denotes its challenging subset, consisting of NAVHARD and NAVSAFE scenes.}
  \label{tab:epdms_all_vertical}
  \centering
  \renewcommand{\arraystretch}{0.8}

  \begin{subtable}[t]{\textwidth}
    \vspace{-0.7em}
    \caption{}
    \label{tab:epdms_v2_all_sub}
    \centering
  \begin{adjustbox}{width=0.9\textwidth}
  \footnotesize
    \begin{tabular}{@{}l |c |c c c c| c c c c c | >{\columncolor{gray!12}}c}
      \toprule
      Method & Modality & \textbf{NC~$\uparrow$} & \textbf{DAC~$\uparrow$} & \textbf{DDC~$\uparrow$} & \textbf{TLC~$\uparrow$} & \textbf{EP~$\uparrow$} & \textbf{TTC~$\uparrow$} & \textbf{LK~$\uparrow$} & \textbf{C~$\uparrow$} & \textbf{EC~$\uparrow$} & \textbf{EPDMS~$\uparrow$} \\
      \midrule
      Human            & --     & 100 & 100 & 99.50 & 100 & 87.51 & 100 & 100 & 98.07 & 90.90 & 89.70 \\
\noalign{\vskip 2pt} \hline\hline\noalign{\vskip 2.5pt} 
Const            & --     & 84.83  & 22.62  & 52.78 & 99.64  & 79.01 & 81.69  & 70.74  & 96.95 & 65.44 & 9.90 \\
      Transfuser~\citep{Chitta2023transfuser} & C \& L & 97.67  & 92.80  & 98.50 & 99.74  & \textbf{87.62} & 96.54  & 95.88  & \underline{98.30} & \textbf{89.05} & 80.40 \\
      DiffusionDrive~\citep{diffusiondrive} & C \& L & \underline{98.21}  & 96.24 & \underline{98.67} & \underline{99.82}  & \underline{87.50} & \underline{97.30}  & \underline{97.02}  & \textbf{98.35} & \underline{88.21} & \underline{83.96} \\
      WoTE~\citep{wote}  & C \& L & \textbf{98.53}  & \underline{96.83}  & 97.90 & \textbf{99.84}  & 86.19 & \textbf{97.88}  & 95.48  & 98.26 & 84.31 & 83.31 \\
\midrule
Ours             & C      & 98.14  & \textbf{97.25}  & \textbf{98.72} & 99.81  & 86.75 & \underline{97.30} & \textbf{97.47}  & 98.28  & 86.07 & \textbf{84.26} \\
      \bottomrule
    \end{tabular}
    \end{adjustbox}
  \end{subtable}

  \vspace{0.75em}

  \begin{subtable}[t]{\textwidth}
    \vspace{-0.5em}
    \caption{}
    \label{tab:epdms_v2_sub}
    \centering
      \begin{adjustbox}{width=0.9\textwidth}
  \footnotesize

    \begin{tabular}{@{}l |c |c c c c| c c c c c | >{\columncolor{gray!12}}c}
      \toprule
      Method & Modality & \textbf{NC~$\uparrow$} & \textbf{DAC~$\uparrow$} & \textbf{DDC~$\uparrow$} & \textbf{TLC~$\uparrow$} & \textbf{EP~$\uparrow$} & \textbf{TTC~$\uparrow$} & \textbf{LK~$\uparrow$} & \textbf{C~$\uparrow$} & \textbf{EC~$\uparrow$} & \textbf{EPDMS~$\uparrow$} \\
      \midrule
      Human            & --     & 100   & 100   & 99.28 & 100   & 85.68 & 100   & 100   & 97.31 & 87.12 & 89.74 \\
      \noalign{\vskip 2pt}\hline\hline\noalign{\vskip 2.5pt} 
      Const            & --     & 84.83 & 22.62 & 52.78 & \underline{99.64} & 79.01 & 81.69 & 70.74 & 96.95 & 65.44 & 9.90 \\
      Transfuser~\citep{Chitta2023transfuser} & C \& L & 94.88 & 68.76 & \underline{96.68} & 96.68 & \underline{84.94} & 92.82 & 91.20 & \textbf{97.49} & \underline{82.62} & 55.28 \\
      DiffusionDrive~\citep{diffusiondrive} & C \& L & 95.78 & 78.99 & 96.41 & \underline{99.64} & \textbf{85.09} & 93.00 & \textbf{96.59} & \textbf{97.49} & \textbf{83.84} & 66.24 \\
      WoTE~\citep{wote} & C \& L & \underline{97.04} & \underline{82.23} & 95.69 & \textbf{99.82} & 81.76 & \textbf{96.23} & 94.43 & \textbf{97.49} & 75.87 & \underline{67.27} \\
      \midrule
      Ours             & C      & \textbf{97.44} & \textbf{83.75} & \textbf{97.44} & \underline{99.64} & 83.32 & \underline{94.88} & \underline{96.05} & 97.31 & 82.56 & \textbf{69.74} \\
      \bottomrule
    \end{tabular}
  \end{adjustbox}
  \end{subtable}
\vspace{-1.0em}
\end{table*}

\subsection{Experiment Results}
We evaluated our proposed method on the NAVSIM benchmark~\citep{Dauner2024NEURIPS,Cao2025CORL}, comparing it against several state-of-the-art methods.
As presented in \cref{tab:pdms_v1}, which details the PDM score from the NAVSIM~v1 benchmark~\citep{Dauner2024NEURIPS}, our approach significantly outperforms existing camera-only methods~\citep{uniad,paradrive,law}.
Furthermore, our camera-only approach performs comparably with multi-modal systems.
On the Drivable Area Compliance~(DAC), we surpass the performance of systems that combine camera and LiDAR~\citep{Chitta2023transfuser,diffusiondrive,wote}.
This indicates that our method addresses the key factors for safe driving in a sensor-efficient manner.

To examine driving quality across additional factors that matter for real-world trajectory planning, we further benchmark on NAVSIM~v2~\citep{Cao2025CORL}, which covers a broader set of evaluation factors than NAVSIM~v1 in \cref{tab:epdms_v2_all_sub}.
\cref{fig:qualitative_fig} provides visual evidence with representative NAVTEST split and shows safe driving in these scenes.
Given our focus on safety-critical situations, we also report results on the combined NAVHARD and NAVSAFE splits to emphasize performance in particularly challenging and safety-critical scenarios.
In \cref{tab:epdms_v2_sub}, our method especially indicates the strongest performance in road-compliance factors~(e.g., DAC, DDC) and remains comparable elsewhere.
Notably, the advantage becomes more pronounced on the NAVHARD~\&~NAVSAFE splits.
This result shows that our approach becomes particularly strong in challenging and safety-critical scenes due to high aleatoric uncertainty.

\begin{figure*}[h]
\centering
\includegraphics[width=0.925\linewidth]{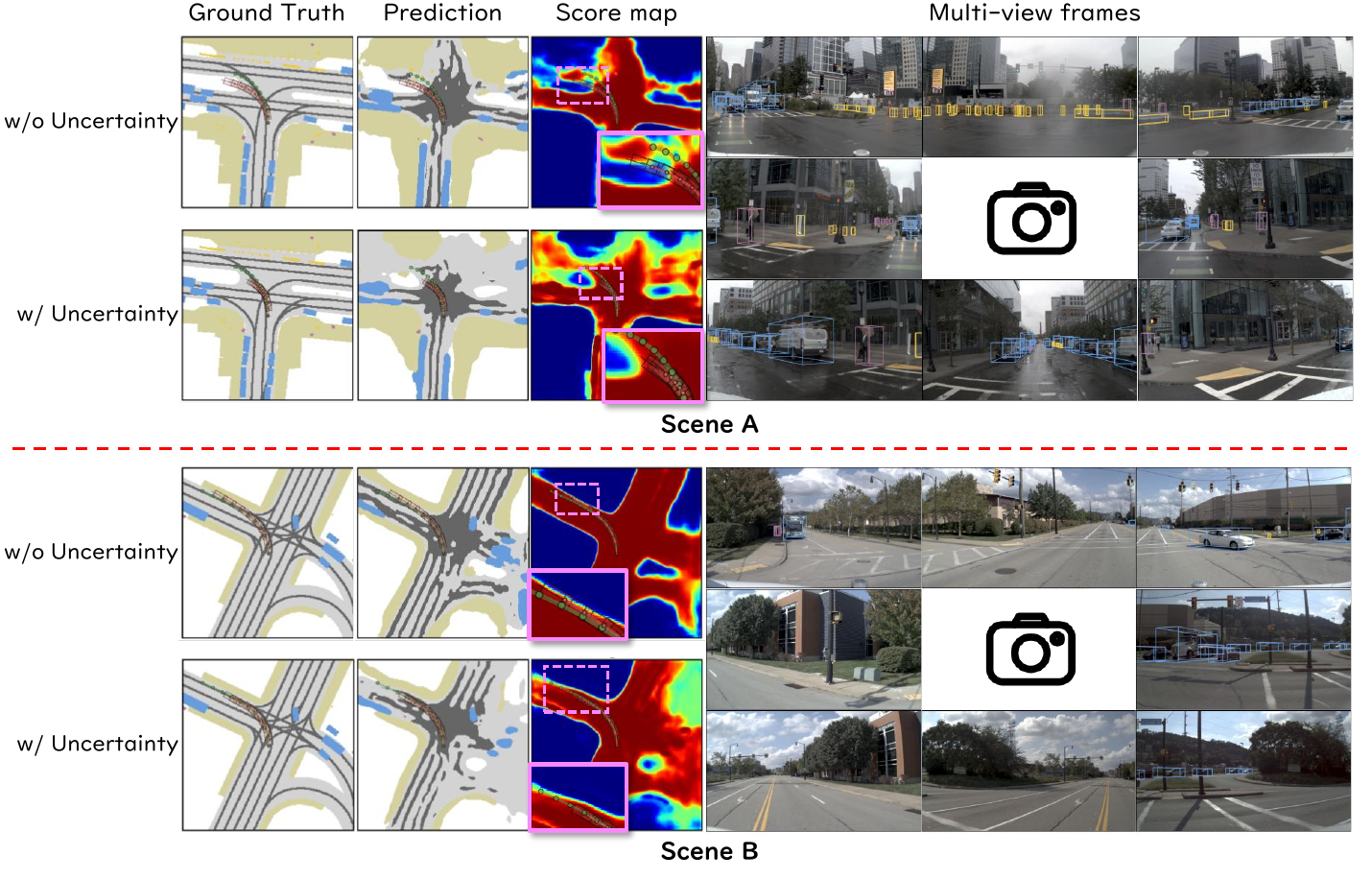}
\vspace{-1.0em}
\caption{Qualitative results of our ablation study on uncertainty. The w/o Uncertainty employs the DiffusionDrive planning module with its perception module replaced by an camera-only BEV encoder. The w/ Uncertainty denotes our proposed method which incorporates uncertainty. The pink bounding boxes highlight how effectively the predicted trajectory leverages uncertainty.
}
\label{fig:ablation_fig}
\vspace{-1.0em}
\end{figure*}
\begin{table}[t]
\vspace{0.2em}
  \caption{Ablations on extended PDM score~(EPDMS; \textbf{Bold}: best; \underline{underline}: second best) with NAVHARD \& NAVSAFE split (C: Camera, L: LiDAR). We highlight the effect of map type (vector vs. segmentation) on performance. This table reports the EPDMS components associated with safety-critical aspects.}

\vspace{-0.5em}
  \begin{adjustbox}{width=\linewidth}
  \renewcommand{\arraystretch}{1.05}
  \large
  \setlength{\heavyrulewidth}{1.5pt}
  \setlength{\lightrulewidth}{0.5pt}
  \centering
  \label{tab:epdms_v2ab_unc}
  \begin{tabular}{@{} l | c | c c c c | >{\columncolor{gray!12}}c}
    \toprule
    Method & Modality & \textbf{NC~$\uparrow$} & \textbf{DAC~$\uparrow$} & \textbf{DDC~$\uparrow$} & \textbf{TLC~$\uparrow$} & \textbf{EPDMS~$\uparrow$} \\
    \midrule
    DiffusionDrive~\citep{diffusiondrive} & C \& L & 95.78  & 79.89  & 96.41 & 99.64  & 66.24 \\
\noalign{\vskip 2pt}\hline\hline\noalign{\vskip 2.5pt}    Our BEV Enc        & C      & 95.60 & 73.43 & 96.14 & \underline{99.64} & 59.70 \\
    Our BEV Enc~\textsubscript{Vector Unc}  & C      & \underline{96.41}  & 73.07  & \textbf{97.94} & \textbf{99.82}  & 59.02 \\
    Our BEV Enc~\textsubscript{Seg Unc}     & C      & 95.78  & \textbf{84.56}  & \underline{97.67} & \underline{99.64}  & \underline{69.19} \\
    \midrule
    Ours             & C      & \textbf{97.44}  & \underline{83.75}  & 97.44 & \underline{99.64}  & \textbf{69.74} \\
    \bottomrule
  \end{tabular}
  \end{adjustbox}
\vspace{-1.0em}
\end{table}
\subsection{Ablation Studies}
This ablation study investigates the individual contributions of our two proposals: uncertainty-aware planning and lane-following regularization, quantifying their effectiveness.

Our model utilizes the planning module from DiffusionDrive~\citep{diffusiondrive}, paired with our custom Transformer-based BEV encoder~(Our BEV Enc) that relies solely on camera input.
We then incrementally employ our uncertainty-aware drivable score map and the lane-following regularization to our model to measure the resulting performance gains.
To validate our approach in complex and safety-critical scenarios, we conducted ablation studies on the challenging NAVHARD and NAVSAFE splits from the NAVTEST split.

Incorporating uncertainty from the segmentation map leads to safer trajectory planning in challenging and safety-critical scenes.
As illustrated in \cref{fig:ablation_fig}, although Scene A shows severe weather degrading perception, planning with uncertainty-aware drivable score map produces considerably risk-aware trajectories than the baseline that relies solely on segmentation map estimation without modeling uncertainty.
Our uncertainty-aware planning method blocks entry into high uncertainty regions and ensures safe trajectories.
In another case shown in \cref{fig:ablation_fig}~Scene B, aleatoric uncertainty induced by long range distance leads to errors near the boundaries of drivable areas.
However, using the uncertainty-aware drivable score map, the planner explicitly avoids such high uncertainty regions and produces safer trajectories.
These qualitative observations clearly demonstrate the benefits of incorporating uncertainty into planning.
Building on this observation, \cref{tab:epdms_v2ab_unc} further compares different uncertainty modeling approaches, showing that utilizing segmentation map uncertainty~(Our BEV Enc~\textsubscript{Vector Unc}) achieves better overall performance than vector map uncertainty~(Our BEV Enc~\textsubscript{Seg Unc}).
These results confirm that perceptual uncertainty is particularly effective in challenging and safety-critical scenarios.

Our second methodology, lane-following regularization, ensures the planning of more robust trajectories.
This effect is indicated in \cref{tab:epdms_v2ab_line_ep_ec}, which reports both the overall EPDMS score and the components related to trajectory robustness.
The model that employs only segmentation map estimation without uncertainty~(Our BEV Enc~\textsubscript{Seg}) serves as comparison model.
When lane-following regularization is applied~(Our BEV Enc~\textsubscript{Seg+Lane}), the trajectories become noticeably more stable and reliable.
This demonstrates the benefit of encouraging the planner to follow the lane centerline during trajectory planning.
In conclusion, combining segmentation map uncertainty with lane-following regularization allows our approach to produce trajectories that are both safer and more robust.

\section{Conclusion}
We address safety-critical challenges in autonomous driving by directly confronting the role of uncertainty in end-to-end planning. Our method explicitly models aleatoric uncertainty in the BEV representation and propagates it to the planner.~This effectively prevents the system from generating unsafe trajectories, such as going off-road.~We also introduce a lane-following regularization that supplies driving norms that are difficult to learn purely from data.~It stabilizes the trajectory, supports consistent lane keeping, and promotes adherence to traffic rules even in uncertain situations. Together, these components yield plans that remain robust, interpretable, and aligned with safe-driving practices across a wide range of conditions.~On the NAVSIM benchmark, our method outperformed existing state-of-the-art works.~In particular, we prove the effectiveness of the proposed method by demonstrating significantly improved performance in safety-critical and challenging scenarios.~This study represents a significant progress toward safe and reliable autonomous driving by effectively handling the inherent uncertainties of real-world environments.~Although our work focuses mainly on uncertainties arising from data and the perception pipeline, future extensions that incorporate uncertainties arising after perception or from the model could make the trajectories even more reliable.
\begin{table}[t!]
\vspace{0.2em}
  \caption{Ablations on extended PDM score~(EPDMS; \textbf{Bold}: best; \underline{underline}: second best) with NAVHARD \& NAVSAFE split (C: Camera, L: LiDAR). We examine the impact of incorporating lane-following regularization. This table focuses on the EPDMS components that specifically measure trajectory robustness.}

  \begin{adjustbox}{width=\linewidth}
  \renewcommand{\arraystretch}{1.05}
  \large
  \setlength{\heavyrulewidth}{1.5pt}
  \setlength{\lightrulewidth}{0.5pt}
  \centering
  \label{tab:epdms_v2ab_line_ep_ec}
  \begin{tabular}{@{} l | c | c c c c c | >{\columncolor{gray!12}}c}
    \toprule
    Method & Modality & \textbf{EP~$\uparrow$} & \textbf{TTC~$\uparrow$} & \textbf{LK~$\uparrow$} & \textbf{C~$\uparrow$} & \textbf{EC~$\uparrow$} & \textbf{EPDMS~$\uparrow$} \\
    \midrule
    DiffusionDrive~\citep{diffusiondrive}    & C \& L & 85.09 & 93.00 & 96.59 & 97.49 & 83.84 & 66.24 \\
    \noalign{\vskip 2pt}\hline\hline\noalign{\vskip 2.5pt} 
    Our BEV Enc        & C      & 84.16 & 93.54 & 94.08 & \underline{97.49} & 75.21 & 59.70 \\
    Our BEV Enc~\textsubscript{Seg}        & C      & \textbf{85.02} & 93.89 & 94.97 & \underline{97.49} & \underline{81.80} & 66.55 \\
    Our BEV Enc~\textsubscript{Seg+Lane}     & C      & \underline{84.78} & \underline{94.08} & \underline{95.87} & \textbf{97.67} & 77.78 & \underline{67.78} \\
    \midrule
    Ours             & C      & 83.32 & \textbf{94.88} & \textbf{96.05} & 97.31 & \textbf{82.59} & \textbf{69.74} \\
    \bottomrule
  \end{tabular}
  \end{adjustbox}
\vspace{-1.0em}
\end{table}

{
    \small
    \bibliographystyle{ieeenat_fullname}
    \bibliography{main}
}

\end{document}